\documentclass[twoside,11pt]{article}

\usepackage{jmlr2e}
\usepackage[algo2e]{algorithm2e}
\usepackage{amsmath}
\usepackage{algorithm}
\usepackage[table]{xcolor}
\usepackage[noend]{algpseudocode}

\makeatletter
\def\BState{\State\hskip-\ALG@thistlm}
\makeatother

\jmlrheading{1}{2021}{1-48}{4/00}{10/00}{pezzotti21a}{Nicola Pezzotti}

\ShortHeadings{MimicBot: Combining Imitation and Reinforcement Learning}{Pezzotti}
\firstpageno{1}

\begin{document}

\title{MimicBot: Combining Imitation and Reinforcement Learning to win in Bot Bowl}

\author{\name Nicola Pezzotti \email nicola.pezzotti@gmail.com \\
       \addr AI, Data Science and Digital Twin Department\\
       Philips Research, Eindhoven, The Netherlands
       \AND
       \addr Department of Mathematics and Computer Science\\
       Eindhoven University of Technology, Eindhoven, The Netherlands}

\editor{Kevin Murphy and Bernhard Sch{\"o}lkopf}
\maketitle

\begin{abstract}
This paper describe an hybrid agent trained to play in Fantasy Football AI which participated in the Bot Bowl III competition.
The agent, MimicBot, is implemented using a specifically designed deep policy network and trained using a combination of imitation and reinforcement learning.
Previous attempts in using a reinforcement learning approach in such context failed for a number of reasons, e.g. due to the intrinsic randomness in the environment and the large and uneven number of actions available, with a curriculum learning approach failing to consistently beat a randomly paying agent. Currently no machine learning approach can beat a scripted bot which makes use of the domain knowledge on the game.
Our solution, thanks to an imitation learning and a hybrid decision-making process, consistently beat such scripted agents.
Moreover we shed lights on how to more efficiently train in a reinforcement learning setting while drastically increasing sample efficiency.
MimicBot is the winner of the Bot Bowl III competition, and it is currently the state-of-the-art solution.
\end{abstract}

\begin{keywords}
  Reinforcement Learning, Imitation Learning, Behavioural Cloning, Deep Learning, FFAI, Bot Bowl
\end{keywords}

\section{Introduction}
Games are the perfect test bench to verify the performance of agent-based machine learning algorithms.
Examples of such approaches can be found in agents that can play ATARI games, Go, DOTA or Starcraft \citep{mnih2015human,silver2017mastering,berner2019dota,vinyals2017starcraft}.
Recently,~\cite{justesen2019blood} presented the Fantasy Football AI (FFAI) framework.
FFAI is based on the Blood  Bowl board game, a fully-observable,  stochastic,  turn-based and modern-style board game played on grid-based game  board.
The framework provides environments to train reinforcement learning algorithms, as well as the implementation of scripted agents that can compete against each others.
Together with the FFAI framework, the authors organized three competitions, called Bot Bowl, where different agents compete against each other to define the winner.

Although the  game seems to be approachable by existing game-playing algorithms, it presents several challenges.
First, all pieces on the board belonging to a player can be moved several times each turn.
Moreover, the number of actions available to each piece can vary considerably, from the ability to attack a single enemy piece, to a movement in an area equal to half the playing board.
This makes the turn-wise branching overwhelming for traditional search-based algorithms.
Scoring  points, corresponding to a touchdown in the game, is  rare  and  difficult.
On top of that, the game has an intrinsic randomness and punishes harshly poor decisions, which act as a confounder for Monte-Carlo based methods.
The combination of the two factors above makes difficult to define heuristics for search algorithms or the application of a reinforcement-learning approach.

In the previous two Bot Bowl completions organized on FFAI, machine-learning bots, were only able to play successfully against a randomly playing agent, not being able to beat a scripted baseline model. The winner in the Bot Bowl competitions relied only on scripted behaviour that encoded domain-expert knowledge.

In this work we present a novel bot, MimicBot, which implement a machine-learning based training strategy which is consistently and largely beating the scripted model provided in the FFAI framework.
MimicBot is the winner of the Bot Bowl III competition, as well as the winner of the award for the most innovative machine learning solution for playing Blood Bowl\footnote{https://njustesen.github.io/ffai/bot-bowl-iii}.
This result is achieved thanks to three contributions of this work:

\begin{enumerate}
    \item We observe that, despite the large decision space, the game is characterized by several symmetries. We propose a deep-learning based actor-critic model that leverages such symmetries using a channel-wise attention mechanism.
    Such model, is small enough to be executed only on a low-end CPU, as required by the Bot Bowl competition, and in stark contrast to the architecture that are often employed in similar settings.
    \item We propose a training strategy that combines imitation- and reinforcement-learning. Such an approach allows to increase the training efficiency by several orders of magnitude. Moreover, thanks to a hybrid training combining the two paradigms, great flexibility is provided in changing the reward for pretrained models, as it counters the so-called catastrophic forgetting that can arise when a shift is introduced in the learned value function.
    \item Finally, we show that a hybrid agent, in which some behaviour is scripted, can be efficiently implemented in the bot. The flexibility of our method, allows to introduce such scripted behaviour from the start, only during inference, or to retrain a bot.
\end{enumerate}

\section{Related work}
In this section we introduce key concept needed to understand our contributions.
We first describe the FFAI framework and the Blood Bowl game.
We then provide an overview of the reinforcement learning problem.
Finally, we introduce imitation learning, with a focus on behavioural cloning.

\subsection{Fantasy Football AI}
Fantasy Football AI is based on Blood Bowl, a board game that is played on a 26x15 square board representing an American football field.
The game has \textbf{perfect information}, it is \textbf{stochastic} and it is\textbf{ turn-based} with \textbf{multiple actions} performed by each player.
Two players play against each other in a 16 turns game, divided in two halves of 8 turns each, with the goal of scoring a \textit{touchdown} in the opposing player \textit{endzone}.
In every turn, a player may activate all his pieces on the board.
Normally, 11 pieces are at the disposal of a player, but the number may reduce during the game due to \textit{injuries} suffered by the pieces.
In the activation, a piece on the board can perform numerous actions based on their status.
The piece can be moved by an amount of squares defined by its movement characteristic.
The piece can \textit{block} an enemy piece in its \textit{tackling zone}. The block consists in rolling one, two or three dices depending on the strength characteristic of the activated and enemy piece, as well as the number of supporting pieces the two players have.
A supporting piece is one that has the enemy piece in its \textit{tackling zone}, without being in the tackling zone of any enemy pieces.
The outcome of the dice roll decides whether the enemy piece is pushed, the enemy or the player piece are going to the ground, meaning that they cannot interact with the enemy until the beginning of the owner's turn. When a player goes to the ground there is a chance that is removed from the board for the rest of the game or until a touchdown is scored, i.e. suffering an injury.
When a piece move outside an enemy tackling zone, it incurs the risk of going to the ground, with the possibility of being removed from the board as for a block.
The risk of going to the ground increases proportionally to the number of enemy tackling zones along the movement path.
An extremely important concept in the game, is the so called \textbf{turnover}.
When a piece fails a critical action, that results in going to the ground or losing the ball, the player's turn immediately ends and the new enemy turn starts.
More detail on the ruleset can be found in~\cite{justesen2019blood}.
Moreover, the game allows for a number of \textbf{re-rolls} which can be used to modify an outcome and avoid a turnover.

The complexity of the game is high compared to Chess or Go. While for Chess the turn-wise branching factor is of $\approx30$ and $\approx300$ for Go,  for Blood Bowl it is $\approx10^{50}$. As a result, the authors of the FFAI framework reports that in 350k games, a random bot could never score a touchdown when playing against another random bot.
The high branching factor makes it particularly challenging to adapt existing algorithms to work in this setting, especially if they are search based.
At the same time, the inability of a random bot to perform meaningfully well, the high degree of randomness and the fact that every mistake is punished harshly by interrupting the player turn, makes it difficult to employ learning paradigms based on trial and error.
Reinforcement Learning using A2C \citep{mnih2016asynchronous} with Curriculum Learning \citep{bengio2009curriculum} was previously applied in the Bot Bowl II competition, but the resulting bot could win only 70\% of the time against a random bot, while it could not beat a scripted bot that has roughly the performance of a novice human player.

In this work we show that, by combining reinforcement learning, imitation learning and a properly designed deep policy network that uses inductive biases, we do not only achieve perfect performance against a randomly playing agent, but we consistently defeat a scripted bot that is estimated to play as well as novice blood bowl players.
Remarkably, our solution can be executed on a laptop computer using only its CPU, and is trained on a desktop computer with a single low end GPU.

\begin{figure}
    \centering
    \includegraphics[width=1.0\linewidth]{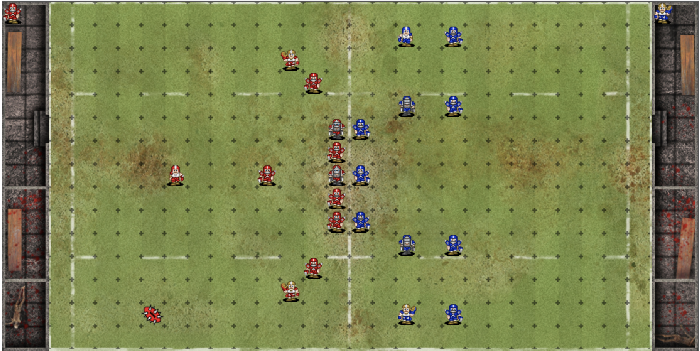}
    \caption{Initial set up of FFAI.}
    \label{fig:my_label}
\end{figure}
\subsection{Reinforcement Learning}
In a Reinforcement Learning (RL) setting, an agent $\mathcal{A}$ interacts with a given environment $\mathcal{E}$ over a number of discrete steps.
At every step $t$, the agent observe a state $\textbf{s}_t$ and, according to a policy $\pi$, selects an action $\textbf{a}_t$ that is to be performed in the environment.
The policy is a mapping function that links states to actions, which in an RL setting, is learned to optimize a scalar reward received by the agent.
The reward is given by the environment to the agent, measuring the quality of the performed action.
This process continues until the environment reaches a terminal state, concluding a so called \textit{episode} in the environment.
We also define the return value at step $t$ as the discounted sum of the rewards
\begin{equation}
R_t = \sum^{\infty}_{k=0}\gamma^k r_{t+k}  ,
\end{equation}
where $\gamma \in \left(0,1\right]$ is the discount factor. The return, encodes the value of the current state and action pair, given the ability to \textit{look-ahead} to the end of the episode.
Overall, we can define the goal of the agent as to maximize the expected return for each state.

Several ways to solve the reinforcement learning problem are possible.
A common way to classify RL algorithms is to distinguish them in methods based on the estimation of \textbf{value functions}, methods that perform a \textbf{policy search} or hybrid methods named \textbf{actor-critic}.

For a comprehensive overview we refer the reader to the work of ~\cite{sutton2018reinforcement} and \cite{arulkumaran2017deep}.
In this work we focus on A2C~\citep{mnih2016asynchronous} as reinforcement learning algorithm of choice, however our solution can be deployed in a wide range of RL settings due to its flexibility.

\subsection{Imitation Learning}
In the most generic sense, imitation learning agents aim at mimicking the human behavior on a given task.
Given the ability to access a human expert acting in the same environment as the agent, a number of states $\textbf{s}$ and the corresponding actions $\hat{\mathbf{a}}$ are captured.
The agent is then trained to mimic the performance of the human given the captured pairs.
This learning paradigm is gaining popularity for a number of reasons.
It enables to train agents on complex tasks without requiring expert knowledge on it. Compared to reinforcement learning, where defining and shaping the reward is extremely important but often difficult to achieve, imitation learning provides a convenient solution as the reward is defined by the actions of the human expert.
This is relevant in a number of fields, such as autonomous driving or  robotics \citep{pomerleau1989alvinn,dosovitskiy2017carla,abbeel2005exploration,duan2017one,codevilla2018end,coates2008learning}.
Moreover, cloud computing made possible to access an increasingly large corpora of expert data behavior.
An example, in the context of gaming, are the available driving simulators or board game implementations available.
Using such examples to develop AI agents is therefore compelling.

Behavioural Cloning is the simplest approach to imitation learning.
The idea behind this approach is to reduce the definition of a policy to a supervised training problem.
We define $\hat{P} = P\left(\textbf{s}|\hat{\pi}\right)$ as the distribution of states \textbf{s} visited by experts, under their expert policy $\hat{\pi}$.
The aim of Behavioural Cloning is to optimize a policy parameterized by $\theta$, in a supervised fashion, for the following objective function:
\begin{equation}
    argmin_{\theta} \mathbb{E}_{(\textbf{s},\hat{\mathbf{a}})\sim \hat{P}}L\left(\hat{\mathbf{a}},\pi_\theta\left(\textbf{s}\right)\right)
\end{equation}
where $(\textbf{s},\hat{\mathbf{a}})$ are state-action pairs obtained from the expert distribution $\hat{\pi}$, $L$ is a loss function that measure how close the action obtained from the learned policy $\pi$ on the current state $\textbf{s}$ matches the action $\hat{\mathbf{a}}$ chosen by the expert.
Based on the task at hand, different choice of $L$ and how to capture the action pairs can be made.

The main strength of Behavioural Cloning lays in its simplicity. However, it has some severe drawbacks, namely the inability to capture drifts in the distributions between the expert policy $\hat{\pi}$ and the learned policy $\pi_\theta$. This is a common problem, as the learned policy will often sample states that are sub-optimal which the human expert will avoid. Therefore, the lack of such states in the training set will hamper the ability to train the agent in a supervised fashion.
More advanced approaches exist in which the drifts between the two policies is avoided by smart sampling and interaction with the experts.
An overview of the most popular imitation learning approaches is provided by~\cite{hussein2017imitation}.

In this work, we demonstrate the potential of using Behavioural Cloning to pre-train a reinforcement learning policy.
Instead of using human experts, we demonstrate that having access to scripted agents enables to achieve superior performance, without the need to put in place large scale collection of human games.
We also show, that combining Reinforcement Learning and Behavioural Cloning enables to avoid the so called \textit{"policy collapse"} that can happen when training with a RL solution.

\section{MimicBot: a hybrid model with channel attention}

In this section, we present the architecture of the deep policy network of MimicBot.
The network is of low computational complexity, as required by the Bot Bowl competition, and requires only few milliseconds to be executed on a low-end CPU.

Before we describe the design choice for the bot, let's introduce the observation space, as well as the action space for the FFAI environment.
The observation space can be divided in two different categories, namely spatial and non-spatial inputs.
The spatial input is a collection of 2D vectors having the shape of the pitch (17x28).
Several channels, more precisely 43, encode information such as the position of the playing pieces, pieces that can be activate during the turn, and the availability of certain skills and action types.
The non-spatial input contains 116 values which encode information about the general state of the game.
This can be the current score, the number of re-roll left, modifiers to the global state of the game which may make more difficult to execute certain actions.
The combination of the two inputs give an input space of size 20584.

The action space can also be seen as a combination of spatial and non-spatial actions.
Concerning the spatial actions, which again are 2D vectors of size 17x28, encodes action to be played on the board. These can represent the movement of a given piece, or the block action against an enemy piece.
The 25 non-spatial actions encodes high-level decisions.
These can be the choice of a given dice roll, or the decision to use a re-roll.

\begin{figure}
    \centering
    \includegraphics[width=0.99\linewidth]{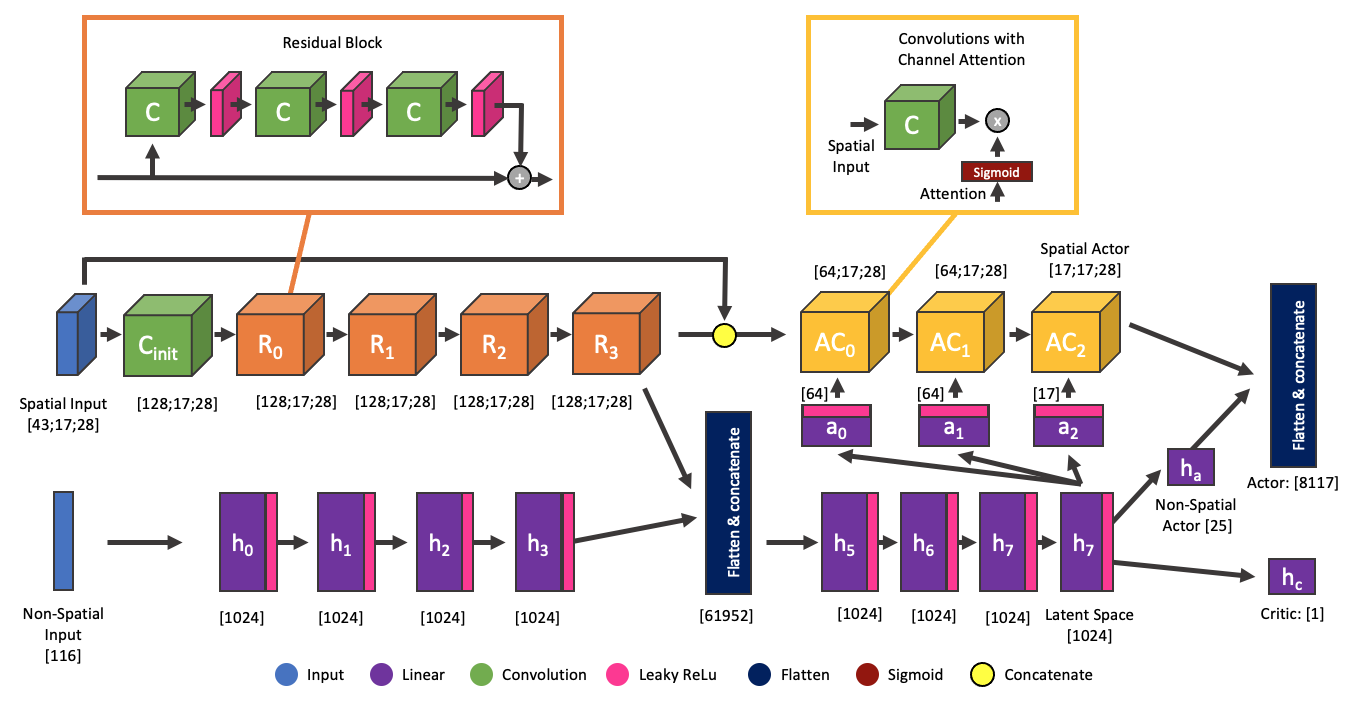}
    \caption{Overview of the MimicBot deep policy network. The spatial decision path preserves spatial correlations in the input (top).
    Information from the non-spatial input is translated to the spatial actions through a channel attention mechanism where a weighting factor is learned.}
    \label{fig:mimicbot}
\end{figure}

The total amount of actions available is 8117.
However, for a given state, only a limited number of actions is available due to the situation on the board.
In MimicBot, we remove impossible actions by setting the network activation to $-\infty$, resulting to a zero probability of being selected after a soft-max operation is applied (see~\cite{kanervisto2020action} for other action space shaping strategies).

Figure~\ref{fig:mimicbot} presents the architecture of the MimicBot policy network.
In the next section we present how the spatial information is processed through the network.
Section~\ref{sec:channel_attention} describes how the non-spatial information is processed through the network and used to influence the computations in the spatial channels.
Finally, Section~\ref{sec:scripted_actions} presents how the bot behavior can be overridden in order to impose some scripted behaviour.

\subsection{Spatial actions}
In previous work, the spatial observations were processed by convolutions, while the non-spatial input was processed by fully connected layers. The two inputs where then combined by flattening the convolutions and by concatenating the result with the output of the fully connected layer.
To reach the final action space of 8117 actions, the concatenated result was then processed by a number of fully connected layers.

In MimicBot, we observe that the design above destroys a relevant inductive bias in the network, more specifically the spatial correlations in the input. This results from the fact that the spatial channels are flattened and processed by the fully-connected layers that do not enforce a preservation of such precious information.
We therefore propose to use a convolutional approach, while we still inject global information from the non-spatial observation using the channel attention mechanism as presented in the next section.
At the top of Figure~\ref{fig:mimicbot}, the processing block of our network is presented.
The input is initially processed by $C_{\text{init}}$, which increase the number of channels to 128.
Four residual blocks \citep{he2016deep} then sequentially process the data. Each block contains three convolutions and 3 leaky-ReLu.
Before further processing is executed, the spatial input is concatenated to the output of the last residual block to retain it in the final decision making steps.
This information is further processed by three convolutions with channel attention, which are described in the next section.
The output of the last convolution with attention is of shape 17x17x28, corresponding to the 17 spatial actions.
This tensor is then flattened and concatenated with the computed non-spatial actions. Note that the inductive bias on the shape of the spatial input is preserved throughout the decision process.

\subsection{Non-spatial actions and channel attention}
\label{sec:channel_attention}
The bottom side of Figure~\ref{fig:mimicbot} presents how the non-spatial input is processed, and how the non-spatial actions are evaluated.
Starting from the non-spatial input, four linear layers followed by leaky ReLu produce 1024 neurons.
Then, the output of the residual blocks, used in the spatial-channels, is flattened and concatenated with output of the last linear layer. This enables to include spatial information in the resulting latent space, which is produced by another four fully connected layers and consisting of 1024 neurons.
This latent space is then used for several purposes.
First, a linear layer uses the latent space to predict the value of the current state. This is the so called Critic in the A2C model.
Then, another linear layer reduces the 1024 neural activations to the 25 available non-spatial actions which compose the final activation space.

Finally the latent space is used to control the so called channel attention mechanism, which is inspired by the attention used in NLP and vision models such as transformers~\citep{vaswani2017attention,devlin2018bert,han2020survey}.
The latent space is converted to a multiplicative effect on the output of the individual channels.
This approach allows to inject global information derived from the latent space, into the bias-preserving convolutional blocks.
Intuitively, it allows to reduce the importance of given spatial actions based on the global context.

\subsection{Scripted actions}
\label{sec:scripted_actions}
Although there are 8117 possible actions in the game, in some context, the decisions to be taken are straightforward.
To this end, it is better to script the behaviour rather than let the bot to predict an action.
As an example, during a block roll, i.e., the attack performed against an enemy piece, the result to be chosen is often clear (always put the enemy to the ground while not incurring in a turnover).
At the same time, removing actions from those available, allows to influence the behaviour of the bot, e.g., by focusing on safer plays or to encode some desired behaviour.
This design choice is known as \textit{action-space shaping}~\citep{kanervisto2020action} and it is widespread in reinforcement learning bots deployed in gaming environments.

In MimicBot, we use several strategies to shape the action space.
For decisions which are straightforward to take, such as which die to chose in a block roll, the policy network is not executed and a scripted routine is used instead.
Similarly, the bot will automatically chose to use a re-roll if, in not doing so, it will cause a turnover, i.e. the loss of the entire turn.
Other activities are automatically scripted for convenience and to accelerate the training of the policy network, e.g., the placing of the ball during kick-off.

A different approach in shaping the action space, is to make certain actions not available to the policy network even though they are legal from a ruleset standpoint.
Removing the actions from the action space is performed similarly to the removal of invalid actions, i.e., by setting the corresponding neuron activation to $-\infty$ before the probability of being chosen is obtained through a soft-max.
With this approach, we can encode different behaviours. For example, our submitted bot to Bot Bowl III, does not allow risky blocks to happen, i.e., by removing them from the available actions when the probability of a turnover is higher than 70\% without using a re-roll.
Another scripted behaviour, is for the ball-carrier, i.e. the piece that has the ball and can score, to be the last to perform a block action.
This is implemented by removing all blocks actions for the ball carrier only if other pieces can perform a block action.
Note that this approach can be implemented and used both during inference and during training. This allows for the training of bots that can be optimized around a scripted behaviour like the one described above.

\section{Training}
In this section, we describe how MimicBot is trained.
With a randomly initialized policy, the chance of performing meaningful actions is almost nonexistent.
To address this issue, a curriculum learning approach based on PPCG~\citep{justesen2018illuminating} was applied to FFAI, but achieved only limited performance against a randomly playing agent.
In this work, we do not employ a curriculum learning strategy but we present an approach based on imitation learning, and more specifically on behavioural cloning, which is described in Section~\ref{sec:bc}.

However, behavioural cloning (BC) cannot exceed the performance of the agent it is imitating, in our case the baseline scripted FFAI agent.
Using the pre-trained BC agent in a reinforcement learning setting leads to a policy collapse.
We address this issue by using a hybrid training strategy, which is described in Section~\ref{sec:rlbc}.
Finally, we adopt a self-playing agent with hybrid decision making to achieve to best performance for our bot.

\subsection{Imitation learning via behavioural cloning}
\label{sec:bc}
The first stage of training MimicBot is performed using a behavioural cloning strategy, as presented in Algorithm~\ref{alg:bc}.
First we collect a number of state-action pairs from an expert policy $\hat{\pi}$, which, in our case, is provided by the scripted bot in the FFAI framework.
We collected 400k state-action pairs $(\hat{\textbf{a}},\textbf{s}) \in I$, by letting two scripted bots play against each other for 200 games.
The set $I$ was then split into a training set $I_{train}$ consisting of 300k pairs, and a validation set of $I_{val}$ with 100k pairs.

\begin{figure}
    \centering
    \includegraphics[width=0.6\linewidth]{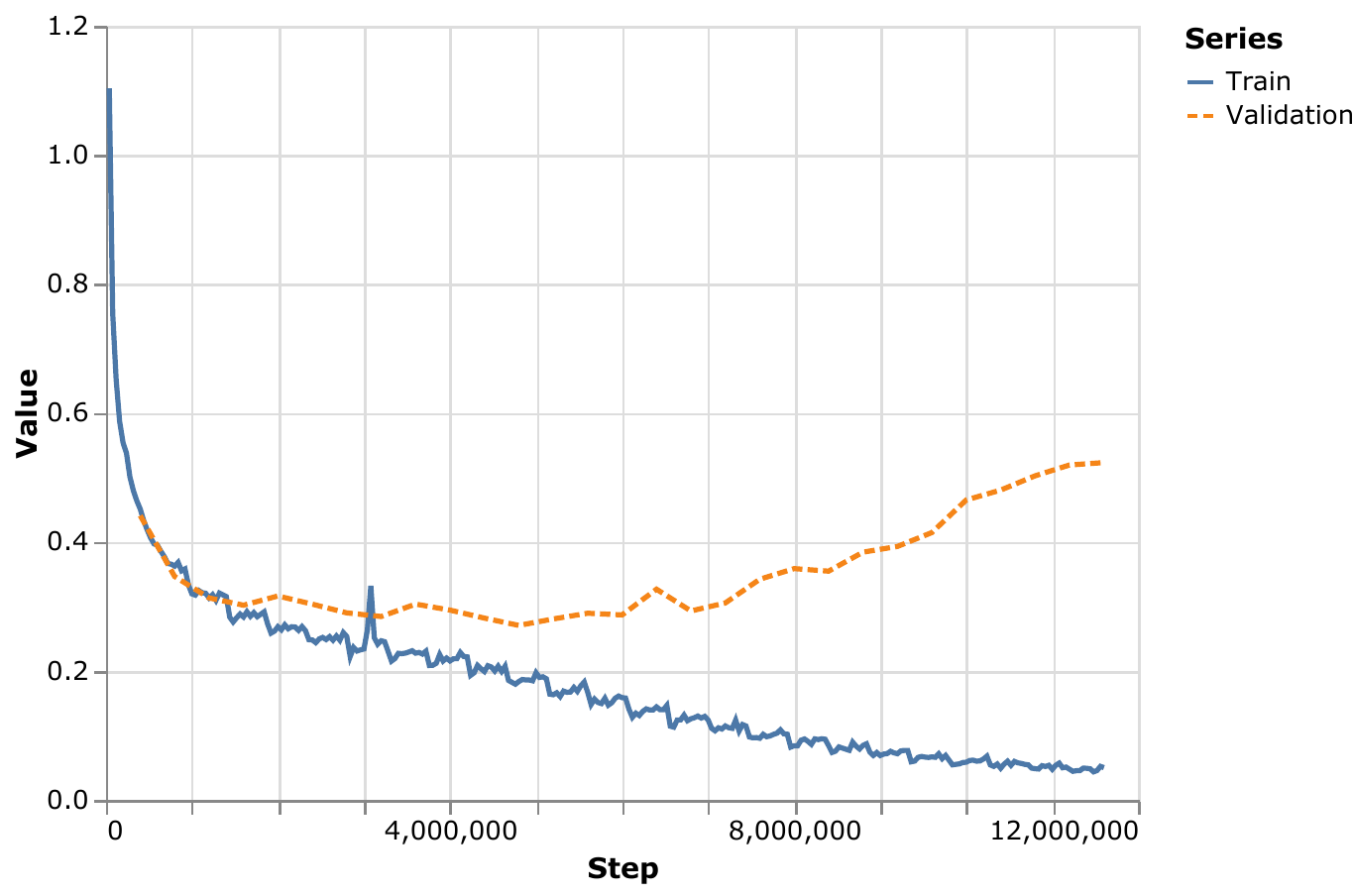}
    \caption{Training of $\pi_\theta^{BC}$ as a classification task. Note that the, despite the poor generalization, it achieves perfect winning rate against a random bot and, although sub-optimally, it can stand against the scripted bot that learnt to imitate.}
    \label{fig:bc_curves}
\end{figure}

A deep policy network $\pi_\theta^{BC}$, as presented in Figure~\ref{fig:mimicbot}, was then trained by treating the action selection as a classification problem.
We train $\pi_\theta^{BC}$ using the negative log-likelihood over the 8117 labels that correspond to the available actions.
Note that, despite the large number of available labels, since we apply the action-space shaping as presented in Section~\ref{sec:scripted_actions}, effectively only a limited number of action in the order of few dozens are to be chosen for any given state.
The network is optimized using the RAdam optimizer \citep{liu2019radam}, a learning-rate of 0.0001, a batch size of 4 and for 30 epochs. The best performance on the validation set is obtained after 12 epochs, as shown in Figure~\ref{fig:bc_curves}. No hyperparameter tuning was performed at this stage.

The model was trained on an RTX 1080 for $\approx$ 48 hours, requiring only 300MB of GPU memory, and reaching peak classification performance after less than a day.
Notably, the model trained in this way achieves perfect results against a randomly playing agent, as it is presented in Section~\ref{sec:results_performance}, beating the RL agent trained with a curriculum learning approach\footnote{See https://njustesen.github.io/ffai/bot-bowl-ii}.
However, as expected it cannot beat the scripted agent from which $I$ was sampled, this is not surprising as $\pi_\theta^{BC}$ only approximates $\hat{a}$ and does nothing to improve over it.

\begin{algorithm}[H]
\textit{// Use behavioural cloning to obtain $\pi_\theta^{BC}$\\}
Given a scripted or human agent $\hat{\pi}$\\
Initialize the imitation buffer $I = []$\\
 \While{$|I| < I_{max}$}{
    \While{initialize the environment $\mathcal{E}$ with two agents $\hat{pi}$}{
        $\hat{\textbf{a}} = \hat{\pi}\left(\textbf{s}\right)$\\
        add $(\hat{\textbf{a}}, \textbf{s})$ in the buffer $I$
    }
 }
 Split $I$ into $I_{train}$ and $I_{val}$\\
 \textit{// Optimize the policy network $\theta^{BC}$, with $L$ being the negative log likelihood, and the actions selection is treated as a classification task}\\
 \While{$\sum_k^{I_{val}} L\left(\hat{\mathbf{a}}_k,\pi^{BC}_\theta\left(\textbf{s}_k\right)\right)$ decreases}{
    $\theta^{BC} := \theta^{BC} - \eta \nabla L\left(\hat{\mathbf{a}}_i,\pi^{BC}_\theta\left(\textbf{s}_i\right)\right)$ with $(\hat{\textbf{a}}_i,\textbf{s}_i) \in I_{train}$
 }

 \caption{\label{alg:bc}Behavioural cloning for training MimicBot}
\end{algorithm}

\subsection{Reinforcement Learning with imitation}
\label{sec:rlbc}
To perform better than the scripted baseline, we need to move beyond the performance achieved through behavioural cloning.
We propose to start from the pretrained policy network $\pi_\theta^{BC}$, and optimize the policy using reinforcement learning, while avoiding a policy collapse using the behavioural cloning objective as presented in the section above.
The resulting policy $\pi_\theta^{MB}$ does not only outperforms the scripted agent, but it can be further improved by using a self-playing mechanism.
In this work we play against the scripted bot and we adopt A2C~\citep{mnih2016asynchronous} as RL algorithm, however, our solution works with other and more advanced algorithms such as TRPO~\citep{schulman2015trust}, PPO~\citep{schulman2017proximal} and several others~\citep{andrychowicz2020matters}.

\begin{figure}
    \centering
    \includegraphics[width=0.99\linewidth]{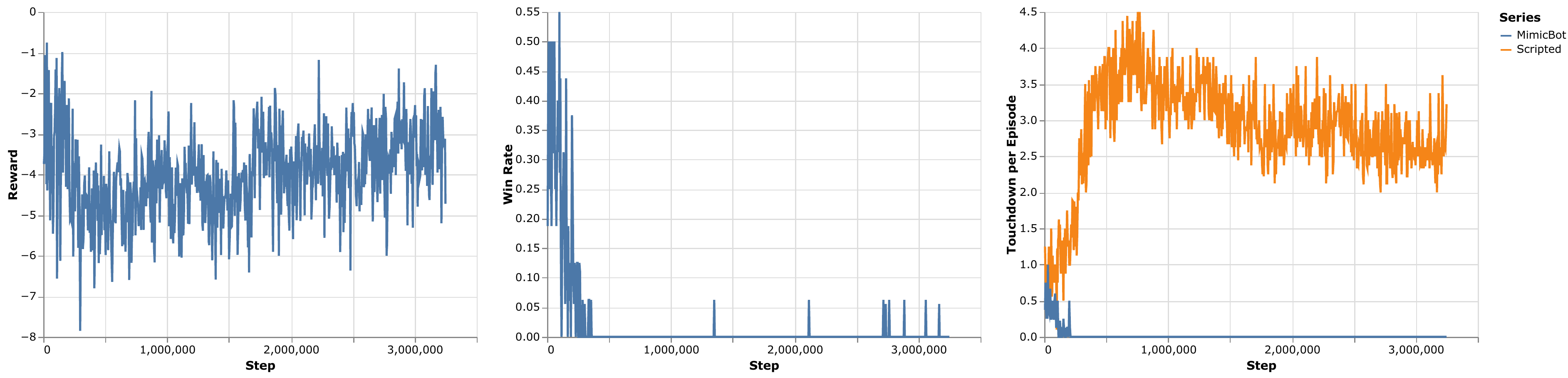}
    \caption{Example of a policy collapse for a model trained with A2C and the pre-trained policy $\pi_\theta^{BC}$.}
    \label{fig:collapse}
\end{figure}

The policy network $\pi_\theta^{BC}$ trained with BC cannot approximate the value function needed in an RL setting, more specifically in an actor-critic solution.
During the first training iterations using A2C, we expect to see the policy learning such value function.
However, the quality of the predicted actions is not constrained in any way, being it derived from the critic component, i.e., the approximated value.
This leads to a policy collapse, as the network is not able to retain the learned behaviour for the actor.
This is a specif case of the catastrophic forgetting~\citep{mccloskey1989catastrophic} that affects neural networks in sequential learning tasks.
Figure~\ref{fig:collapse} show the training curves using A2C starting from the pretrained policy.
It can be seen, that, although initially the bot is performing well, it soon collapse to a bot that can neither win nor score.

We propose a simple and effective way to retain what learned by $\pi_\theta^{BC}$, by mixing the RL objective with the BC objective as presented in Section~\ref{sec:bc}.
Our algorithm works as follows.
We execute the RL algorithm, A2C in this case, and computing the corresponding updates to the parameter $\theta$.
We then optimize $\theta$ in the very same way as for the behavioural cloning training for a number of iterations.
In this way, $\pi_\theta^{MB}$ learns to approximate the value function using A2C, while BC is used to correct and retain the learned behaviour by the critic.

\begin{algorithm}[H]
\textit{// Compute $\pi_\theta^{MB}$ starting from the pretrained policy $\pi_\theta^{BC}$\\}
$\pi_\theta^{MB} \Leftarrow \pi_\theta^{BC}$\\

\While{RL learner updates $\theta$}{
    execute $C_{RL}$ updates of $\theta$ as prescribed by the algorithm of choice\\
    execute $C_{BC}$ updates of $\theta$ using $I_{train}$
}

\caption{\label{alg:rl}RL training with BC correction}
\end{algorithm}

In practice, both the updates for A2C and BC are computed with RAdam and a batch size of 5.
Two hyper-parameters, $C_{BC}$, $C_{RL}$, represent how many gradient updates are performed with a method over the other control how dominating is one update over the other.
The simple strategy as presented in Algorithm~\ref{alg:rl}, can be implemented in the trainer functions, also for distributed environments such as SEED RL~\citep{andrychowicz2020matters}.
At the beginning of the training of the policy $\pi_\theta^{MB}$, they are equal, meaning that the two methods have the same importance over the learned policy.
In our case we, we set $C_{BC} = C_{RL} = 40$, such that, with a batch size of 5, 200 state-action pairs are used for each method.
Changing the two hyper-parameters allows for controlling how much the bot sticks to the BC policy and how much it drifts away towards the behaviour purely learned using A2C.
After 12M of steps executed using A2C, we set $C_{BC}=0$, meaning that no BC iterations are performed.

\begin{figure}
    \centering
    \includegraphics[width=0.9\linewidth]{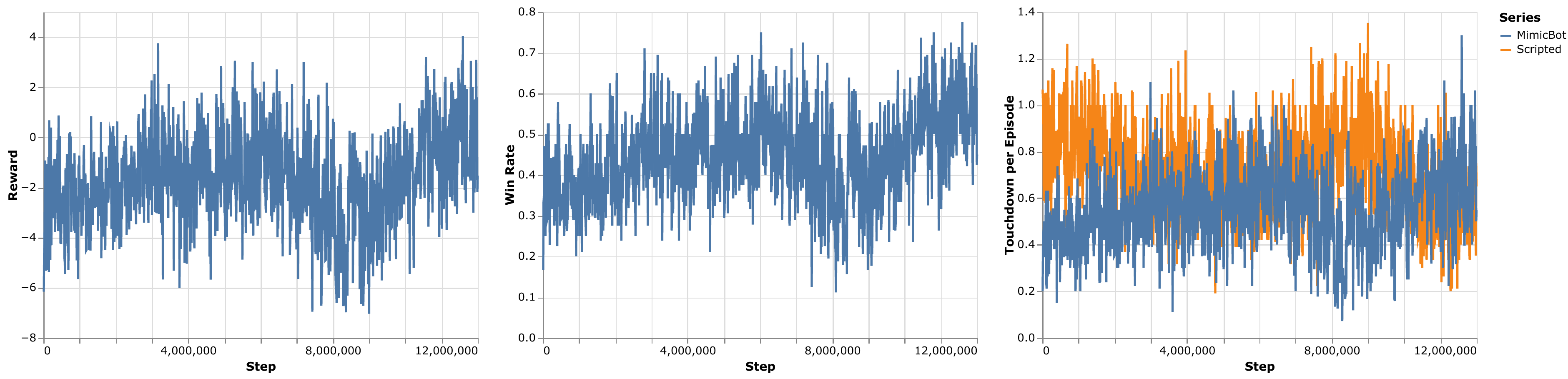}
    \caption{Training of $\pi_\theta^{MB}$, with $C_{RL}=40$ and $C_{BC}=40$ .}
    \label{fig:training_RLBC}
\end{figure}

\begin{figure}
    \centering
    \includegraphics[width=0.9\linewidth]{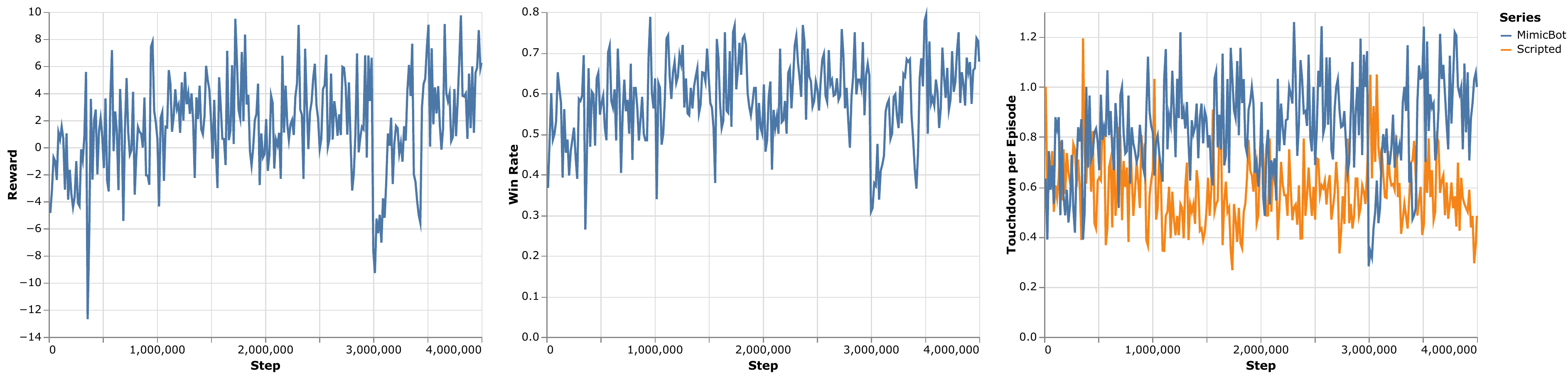}
    \caption{Training of $\pi_\theta^{MB}$, with $C_{RL}=40$ and $C_{BC}=0$ .}
    \label{fig:training_RL}
\end{figure}

Figure~\ref{fig:training_RLBC} shows the training curves for MimicBot when A2C is combined with BC. It can be seen that, compared to Figure~\ref{fig:collapse}, the bot retains the initial performance and actually improves the winning rate over time.
After 12M game steps, we remove the BC correction making the policy free to deviate from the behaviour which was constrained by the scripted bot.
Figure~\ref{fig:training_RL} show the resulting curves, where we can see that the agent consistently outperforms the scripted bot.
Note that the only hyper-parameter exploration we performed was on the learning rate for the updates with A2C, where we tried to set the learning rate to $10^{-6}$, $5\times10^{-6}$, $10^{-6}$, with $5\times10^{-6}$ providing the best performance.

Finally, to further refine the policy we train MimicBot using a self-playing strategy for a limited number of game states.
More precisely, we let the agent play against 10 enemies - 8 of them adopt the policy $\pi_\theta^{MB}$, while 2 still play against the scripted bot.
The policy $\pi_\theta^{MB}$ is updated for the enemies every 500 game states, and we train for 1.5M states.
In the next section, results of our model are presented.

\section{Results}
We test several stages of MimicBot against each other and against the random baseline and the scripted bot which simulates a novice human player of Blood Bowl.
Note that, the best machine learning bot playing FFAI, could not achieve perfect performance against a randomly playing agent, while could not win a single game against the scripted bot\footnote{See the results of the Bot Bowl II competion at https://njustesen.github.io/ffai/bot-bowl-ii and the description of the bot at https://github.com/mrbermell/ffai/tree/GbgBot-master}.
For this reason, testing against the scripted bot allow to verify if state-of-the-art performance is achieved.

\subsection{Performance analysis}
\label{sec:results_performance}

Table~\ref{tab:results} presents how many of 300 games are won, drawn or lost, for all the different combinations of bots.
The bots are \textbf{Random}, which randomly selects a valid action, \textbf{Scripted} which is based on domain knowledge and simulate a novice human player, \textbf{BC}, which is trained using behavioural cloning as in Section~\ref{sec:bc}, \textbf{MimicBot}, which is trained using A2C and the hybrid training as presented in Section~\ref{sec:rlbc}, \textbf{MB Self-Play}, which extends the previous bot using self play.
Finally, the self-playing bot is extend with \textbf{Hybrid} strategies, \textbf{Basic} and \textbf{Advanced}, where scripted actions are selected as described in Section~\ref{sec:scripted_actions}.
\begin{table}[ht]
\centering
\resizebox{\textwidth}{!}{%
  \begin{tabular}{ | l | c | c | c | c | c | c | c |}
    \hline
    \textbf{Win/Draw/Loss}  & Random & Scripted  & BC & MimicBot & MB Self-Play & Hybrid Basic & Hybrid Adv. \\ \hline
    Random & \cellcolor{gray!25} 0/300/0 & \cellcolor{red!25}0/0/300 &    \cellcolor{red!25}0/0/300 & \cellcolor{red!25}0/0/300 & \cellcolor{red!25}0/0/300 & \cellcolor{red!25}0/0/300 & \cellcolor{red!25}0/0/300 \\ \hline
    Scripted & \cellcolor{green!25}300/0/0 & \cellcolor{gray!25} 69/119/112 &   \cellcolor{green!25}140/99/61 & \cellcolor{red!25}98/101/101 & \cellcolor{red!25}82/102/116 & \cellcolor{red!25}45/74/181 & \cellcolor{red!25}35/77/188 \\ \hline

    BC &\cellcolor{green!25} 300/0/0 & \cellcolor{red!25}61/99/140 & \cellcolor{gray!25}102/96/102 & \cellcolor{red!25}51/75/174 & \cellcolor{red!25}21/66/213 & \cellcolor{red!25}12/28/260 & \cellcolor{red!25}14/32/254 \\ \hline
    MimicBot  & \cellcolor{green!25}300/0/0 & \cellcolor{green!25}114/97/89 &   \cellcolor{green!25}174/75/51 &\cellcolor{gray!25} 104/106/90& \cellcolor{red!25}73/90/137 & \cellcolor{red!25}37/76/187 & \cellcolor{red!25}48/83/169 \\ \hline
    MB Self-Play & \cellcolor{green!25}299/1/0 & \cellcolor{green!25}126/97/77  &\cellcolor{green!25} 213/66/21 &\cellcolor{green!25} 137/90/73 & \cellcolor{gray!25} 113/86/101& \cellcolor{red!25}55/90/155 & \cellcolor{red!25}64/77/159 \\ \hline
    Hybrid Basic & \cellcolor{green!25}300/0/0 & \cellcolor{green!25}181/74/45  & \cellcolor{green!25}260/28/12 & \cellcolor{green!25}187/76/37 & \cellcolor{green!25}155/90/55 &  \cellcolor{gray!25} 108/89/103 & \cellcolor{red!25}104/90/106 \\ \hline
    Hybrid Adv. & \cellcolor{green!25}300/0/0 & \cellcolor{green!25}188/77/35  & \cellcolor{green!25}254/32/14 & \cellcolor{green!25}169/83/48 &  \cellcolor{green!25}159/77/64 &\cellcolor{green!25} 106/90/104 &  \cellcolor{gray!25} 116/96/88 \\
    \hline
  \end{tabular}}
\caption{\label{tab:results} Results for different bots playing in 300 games.}
\end{table}

First, we observe that the games on the diagonal of the table present quite some variance, showing the influence of the stochastic element that is intrinsic in the game.
All bots achieve perfect performance against the randomly playing bot, with the exception of 1 game that is lost by the self-playing MimicBot.
Notably, the BC bot, performs perfectly against the random agent, in stark contrast with previously developed solutions that achieves limited performance and requires many days of training.
The  BC bot, although it can win many games against the scripted agent that is trained to replicate, it cannot beat it consistently.
However, after it is trained using the hybrid strategy as presented in Section~\ref{sec:rlbc}, it consistently beats it and even more so when the self-playing strategy is used.
The best results are obtained when the self-playing bot, is enriched with the hybrid decision making as described in Section~\ref{sec:scripted_actions}.
The advanced bot, wins 62\% of the game, while it loses only 11\% against the scripted agent.
These results are corroborated by the touchdowns scored and suffered as presented in Table~\ref{tab:results-td}.
Moreover, note that for the bots presented here, only limited hyper-parameter tuning and training epochs were performed and we expect that better performance can be achieve when the training is scaled up substantially.

\begin{table}[ht]
\centering
\resizebox{\textwidth}{!}{%
  \begin{tabular}{ | l | c | c | c | c | c | c | c | c |}
    \hline
    \textbf{TD Scored/ TD Suffered} & Random & Scripted  & BC & MimicBot & MB Self-Play & Hybrid Basic & Hybrid Adv.\\ \hline
    Random & \cellcolor{gray!25} 0/0 & \cellcolor{red!25}0/3.95 &   \cellcolor{red!25}0/3.27 & \cellcolor{red!25}0/3.15 & \cellcolor{red!25}0/3.18 & \cellcolor{red!25}0/3.92 & \cellcolor{red!25}0/3.96 \\ \hline
    Scripted & \cellcolor{green!25}3.95/0 & \cellcolor{gray!25} 0.53/0.75 &  \cellcolor{green!25}0.89/0.48 & \cellcolor{red!25}0.72/0.70 & \cellcolor{red!25}0.73/0.86 & \cellcolor{red!25}0.47/1.22 & \cellcolor{red!25}0.45/1.31 \\ \hline
    BC & \cellcolor{green!25}3.27/0 & \cellcolor{red!25}0.48/0.89 & \cellcolor{gray!25} 0.68/0.68 & \cellcolor{red!25}0.41/1.02 & \cellcolor{red!25}0.28/1.27 & \cellcolor{red!25}0.13/1.65 & \cellcolor{red!25}0.19/1.69 \\ \hline
    MimicBot & \cellcolor{green!25}3.15/0 & \cellcolor{green!25}0.78/0.71 &  \cellcolor{green!25}1.02/0.40 & \cellcolor{gray!25} 0.65/0.60 & \cellcolor{red!25}0.59/0.92 & \cellcolor{red!25}0.38/1.25 & \cellcolor{red!25}0.39/1.07 \\ \hline
    MB Self-Play & \cellcolor{green!25}3.18/0 & \cellcolor{green!25}0.96/0.73 &\cellcolor{green!25} 1.27/0.28 &\cellcolor{green!25} 0.92/0.59  &  \cellcolor{gray!25} 0.79/0.75 & \cellcolor{red!25}0.54/1.15 &\cellcolor{red!25}0.57/1.13  \\ \hline
    Hybrid Basic & \cellcolor{green!25}3.92/0 & \cellcolor{green!25}1.22/0.47 &  \cellcolor{green!25}1.65/0.13 & \cellcolor{green!25}1.25/0.38 & \cellcolor{green!25}1.15/0.54 &  \cellcolor{gray!25} 0.86/0.81 & \cellcolor{red!25}0.74/0.78 \\ \hline
    Hybrid Adv. & \cellcolor{green!25}3.96/0 & \cellcolor{green!25}1.31/0.45 &  \cellcolor{green!25}1.69/0.19 & \cellcolor{green!25}1.07/0.39 & \cellcolor{green!25}1.13/0.57 &\cellcolor{green!25} 0.78/0.74 &  \cellcolor{gray!25} 0.83/0.72 \\
    \hline
  \end{tabular}}
\caption{\label{tab:results-td} Number of touchdowns scored by the two bots in the games of Table~\ref{tab:results}}
\end{table}
\subsection{Model analysis}
\label{sec:model_analysis}
\begin{figure}
    \centering
    \includegraphics[width=0.99\linewidth]{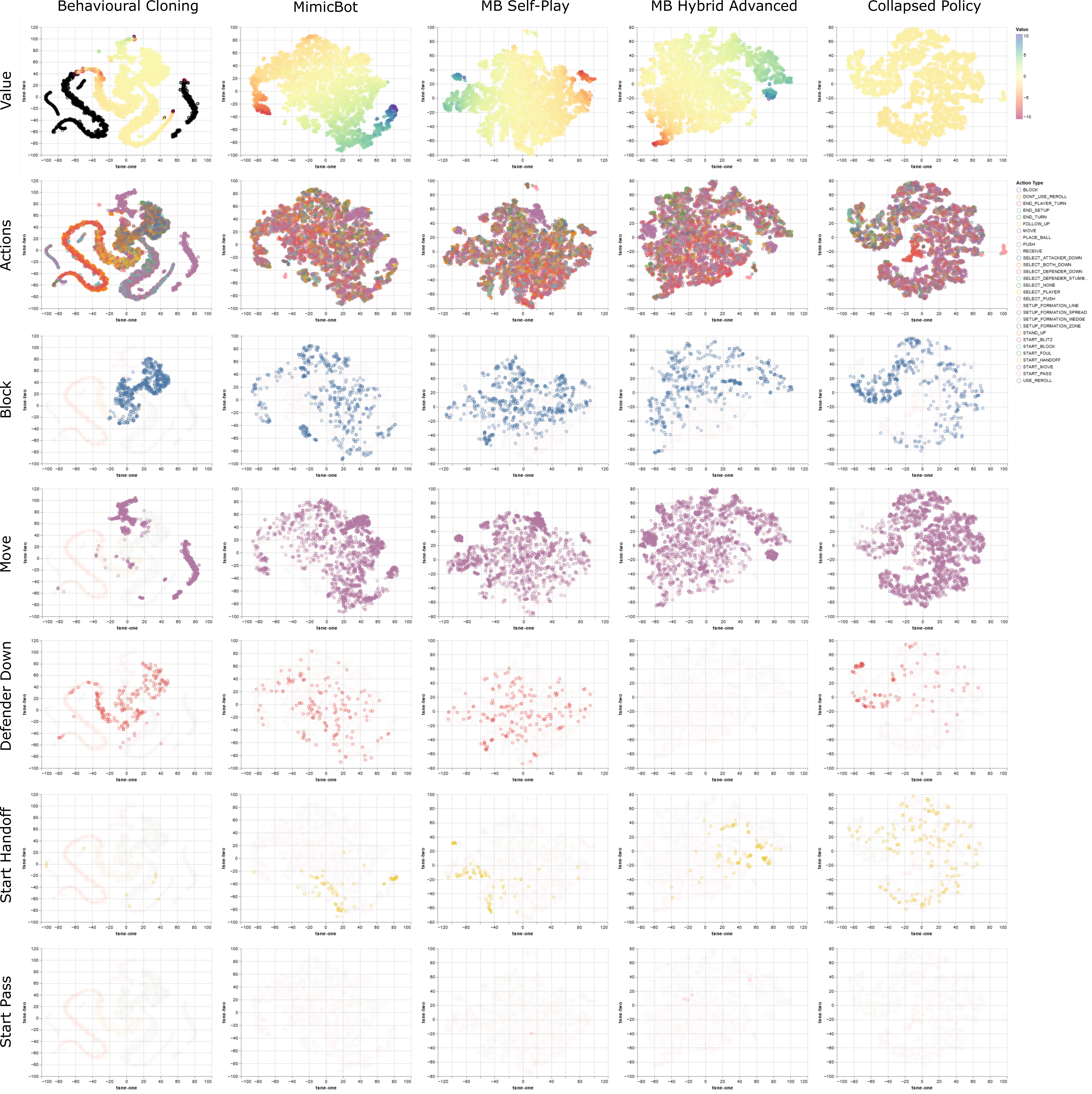}
    \caption{Analysis of the latent space for 5 of the bots presented in Section~\ref{sec:results_performance}. Each point represents a game state obtained by playing against the scripted bot 20 times. A detailed description is available in Section~\ref{sec:model_analysis}.}
    \label{fig:latent_analysis}
\end{figure}

To validate our hypothesis, we conduct an analysis through dimensionality reduction of the latent space of our deep policy network.
This is a common approach to better understand how different game states are perceived by the agent and how they relate to the estimated value function~\citep{mnih2015human,pezzotti2017deepeyes}.
Figure~\ref{fig:latent_analysis}, presents tSNE embeddings~\citep{van2008visualizing} computed with the Approximated-tSNE method~\citep{pezzotti2016approximated}, of the game states obtained by letting different bots playing against the scripted agent for 20 games.

The first row present the embedding, where each game state $\textbf{s}$ is represented as a point and color coded according to the predicted value, while the second row shows the action type that the agent decided to perform.
The remaining rows, shows specific types of actions and their placement in the embedding. Note that, if two states are close in the embedding, they are perceived as similar by the policy network and the selected action will also be similar.

Some interesting insights are immediately visible.
The BC agent, does not learn a meaningful value function, as expected as not explicitly modeled in the training.
Moreover, it perceives different types of actions in its latent space. For example, the \textit{Block} action, in blue, is clearly separated by the \textit{Move} action in purple.
This fact demonstrated that the agent is not able to distinguish favorable board positions, while it rather remembers states that are then mapped to the action labels.

For the MimicBot embeddings, the results are drastically different. The model learnt to estimate the value function, identifying high value states, where a touchdown is about to be scored, from low ones.
Notably, the action types are now distributed along the data manifold, showing that the agent is self-organizing along the notion of value in the game.
Note that, for the hybrid model with advanced scripted actions, no \textit{Defender Down} action is present. This is caused by the fact that this specific action is scripted, and does not require an execution of the policy. This in turn makes the decision space smaller, and the model more efficient.

Another relevant insight is given by the \textit{Start Handoff} and \textit{Start Pass} actions.
An Handoff is a rather safe action, in which a piece with the ball passes it to another piece, which can then move much further. In this case the pieces, have to be close on the board and only one handoff can be performed in a turn.
A slightly more risky action is a pass, which works similarly to the handoff, but he ball can be passed even if the two pieces are not close on the board. The distance between the pieces decides the difficulty of the roll.
We can observe that the BC only performs a limited number of handoffs. This shows a lack of understanding of the importance of the move, which is then recovered by the MimicBot agents.
More interestingly, the self-playing and hybrid bots learned to make use of the pass action.
This is quite interesting, as the scripted bot did not play any pass action. This shows that the agent found a strategy that was not coded in the original bot, demonstrating the ability to explore the state-action space despite the imitation learning approach.

Finally, the last column show the model trained with A2C after a BC pre-training. We see that the model is actually better organizing the latent space, similarly to what is seen in the MimicBot embeddings. However, the value function is collapsed to 0, and the agent cannot properly organize the actions according to the notion of value as seen in Figure~\ref{fig:collapse}.

\section{Discussion}

MimicBot achieves state-of-the-art performance on FFAI thanks to several innovations.
First, inspired by recent work on Geometric Deep-Learning~\citep{bronstein2021geometric}, we observe that the game is characterized by local equivariances and global invariants.
An example of equivariance is the block action and how the local support from friendly and enemy pieces is given.
We designed the policy network to support this effectively, by making use of a specific spatial processing path, which is then controlled by a channel attention mechanism.
Additional future work may comprise the addition of equivariant convolutions in order to fully leverage local symmetries~\citep{romero2020attentive}.
The engineering of new agents for given environments can greatly benefit from the understanding of such symmetries, which can then leveraged in the architecture design. Such design decisions are at the core of several award winning agents~\citep{berner2019dota,vinyals2017starcraft}.

Another research direction that we integrated in MimicBot concerns hybrid deep learning models.
Such models integrate strong inductive biases to boost performance of deep learning solutions, examples of which can be found in several fields, from computer vision to healthcare ~\citep{ardabili2019advances,yang2018proximal,pezzotti2020adaptive}.
Note that the hybrid agent is implemented by overriding the decisions that would be made under the current policy. We expect that, when the hybrid decisions are implemented during training, a better performing can be obtained as the agent during training will have better performance and no distribution shift will be present during inference.

Besides the agent design, our main contribution has to do with the combination of behavioural cloning and reinforcement learning.
We demonstrate that behavioural cloning, with its simplicity, is a valid choice to improve the training of a reinforcement learning agent.
First, we showed that it acts as a powerful way to pre-train the agents. To this end, if scripted game engines are available, it is beneficial from a sample efficiency perspective to pre-train the agent with behavioural cloning.
We also showed, that behavioural cloning is not sufficient to effectively pre-train the RL agent, but it has to be deployed together with ways to avoid policy collapse.
We present a simple and effective solution and we demonstrated it on our game.
Note that this is easily integrated in most RL algorithms, including highly parallelizable training approaches.
We also demonstrated that this approach allows to train well performing agents under resource constrains, in contrast to the trend in the field to favour high distribution to scale up the training.

At the same time, we did not analyze the impact of our approach w.r.t. the exploration and exploitation trade-off. For example, it is unclear at this stage, how different the learnt policy is from the initial scripted agent.
Some differences however are visible, as demonstrated in Section~\ref{sec:model_analysis}, where we see that the agent consistently selects an action that was not performed by the scripted agent.
For example, MimicBot learns to defend the field by keeping piece not directly engaged with the enemy and by stacking tackle zones so that enemy pieces cannot reach the touchdown zone. This is a behaviour that was not present in the original bot, but we cannot conclude that all the optimal states can be reached.
To further understand the agent, we want to extend the analysis presented in Section~\ref{sec:model_analysis}, with more states and hierarchical embedding representations~\citep{pezzotti2016hierarchical}.

Another practical advantage of our solution is that allows for efficient experimentation on reward shaping.
During the development of an agent, the designer can define a shaped reward such that the goals in the environment are more easily satisfied.
However, reward shaping, is a trial-and-error iterative process~\citep{ng1999policy}.
When the shape of the reward is changed, say to remove shaped rewards that are introduced just to accelerate the learning, the value function is changed and the policy risk to collapse.
By adopting our hybrid BC-RL training policy, one can avoid such collapse by training an agent with BC, and 'protecting' the old policy while the new value function is learned. This is a practical and easy to implement approach that allow for faster design iterations.
Although we believe that this holds true for several environments, this has to be tested on a diverse range of cases and policies.

\section{Conclusions and Future Work}
In this work we achieve state-of-the-art results using a machine learning model in the FFAI framework.
This is demonstrated by the victory of MimicBot in Bot Bowl III, both in the tournament and by being the recipient of the award for the most innovative machine learning bot.
Key to our performance, is a policy network that is designed to make use of symmetries in the game and imposes strong inductive-biases in the computations.
However, the design of the network alone is not sufficient.
Given the complexity and randomness in the game, reinforcement learning (RL) agents are extremely sample inefficient, even when a curriculum learning approach is adopted.
We propose to adopt a behavioural cloning (BC) training as a classification task. We then propose a hybrid imitation and reinforcement learning setting to avoid policy collapse.
In the future, we want to investigate how BC in FFAI can benefit from having access to human played games~\footnote{See for example https://en.wikipedia.org/wiki/Blood\_Bowl\_2}, and how longer and more distributed training can improve the performance of MimicBot.
Moreover, we want to explore how our solution is beneficial in different environments and for different RL algorithms.


\acks{Thanks to Niels Justesen and Mattias Bermell Rudfeldt for the discussions and Niels Justesen and the other FFAI creators for the support in the creation of MimicBot.}


\vskip 0.2in
\bibliography{sample}

\end{document}